\documentclass{article}

\usepackage{arxiv}

\usepackage{authblk}
\usepackage{graphicx, svg}
\usepackage{caption}
\usepackage{subfigure, subcaption}
\usepackage{array, multicol, multirow, booktabs}
\usepackage{footnote}
\usepackage{amsmath, amssymb}
\makesavenoteenv{table}
\usepackage{color}
\usepackage{soul}
\usepackage{float}
\usepackage{etoolbox}
\usepackage{wrapfig}

\usepackage[breaklinks]{hyperref}

\usepackage[dvipsnames]{xcolor}
\usepackage[normalem]{ulem}

\makeatother

\newcolumntype{M}[1]{>{\centering\arraybackslash}m{#1}}

\date{}
\begin{document}
    % \begin{frontmatter}
        \title{Revisiting Change Detection Methods for their Application to Serac Fall Time-Lapse Monitoring}
        
        \author[1, 2]{Arthur Dérédel\thanks{\texttt{arthur.deredel@liris.cnrs.fr}}}
        \author[1]{Carlos Crispim-Junior}
        \author[2]{Pierre Lemaire}
        \author[2]{Johan Berthet}
        \author[1]{Laure Tougne Rodet}
        
        \affil[1]{Université Lumière Lyon 2, CNRS, Ecole Centrale de Lyon, INSA Lyon, Université Claude Bernard Lyon 1, LIRIS, UMR5205\\ 69676, Bron, France\\}
        \affil[2]{Styx4D, 19 rue lac Saint André, Le Bourget-du-Lac, 73370, France}
        
        \maketitle
        
        \begin{abstract}
            In an era where climate change aggravates environmental uncertainties, the identification and detection of event precursors are becoming crucial to mitigate the impacts of disastrous natural hazards. While classical sensors such as interferometric lasers or seismometers are reliable, their widespread deployment is often hindered by logistical and economic barriers, leaving numerous blind spots. Time-lapse cameras, which already provide cost-effective, high-resolution visual context to such sensors, present a promising alternative. However, processing their output automatically faces significant challenges, notably linked to extreme shape and lighting variations. Overcoming those issues is essential to deploy them at large-scale as a monitoring tool.
            This paper introduces a novel sub-task of change detection, namely volumetric change detection, applied to time-lapse cameras and slope instabilities. We conduct a comprehensive review of state-of-the-art change detection methods and related tasks, analyze their core components and assess their applicability to this context. To that end, we introduce the new dataset SeracFallDet, which contains serac fall annotations and has been thoroughly annotated to meet the latter demand. Through generalization experiments, we demonstrate that dense and semi-dense feature matching, although not trained specifically for this task, exhibit robust performance. Alternatively, supervised approaches struggle with data scarcity and annotation imbalance. This suggests that hybrid methods may offer a path forward by leveraging the strengths of both tasks. These findings highlight the potential of feature matching techniques and the need for further innovation to overcome the challenges of real-world deployment in environmental monitoring.
        \end{abstract}
        
    %% Keywords
    % \begin{keyword}
    % Computer vision \sep Environmental monitoring \sep Volumetric change detection \sep Time-lapse camera 
    % \end{keyword}
        
    % \end{frontmatter}
    
    \section{Introduction}

\begin{figure*}[ht]
            \centering
            \includegraphics[width = 1.0\textwidth]{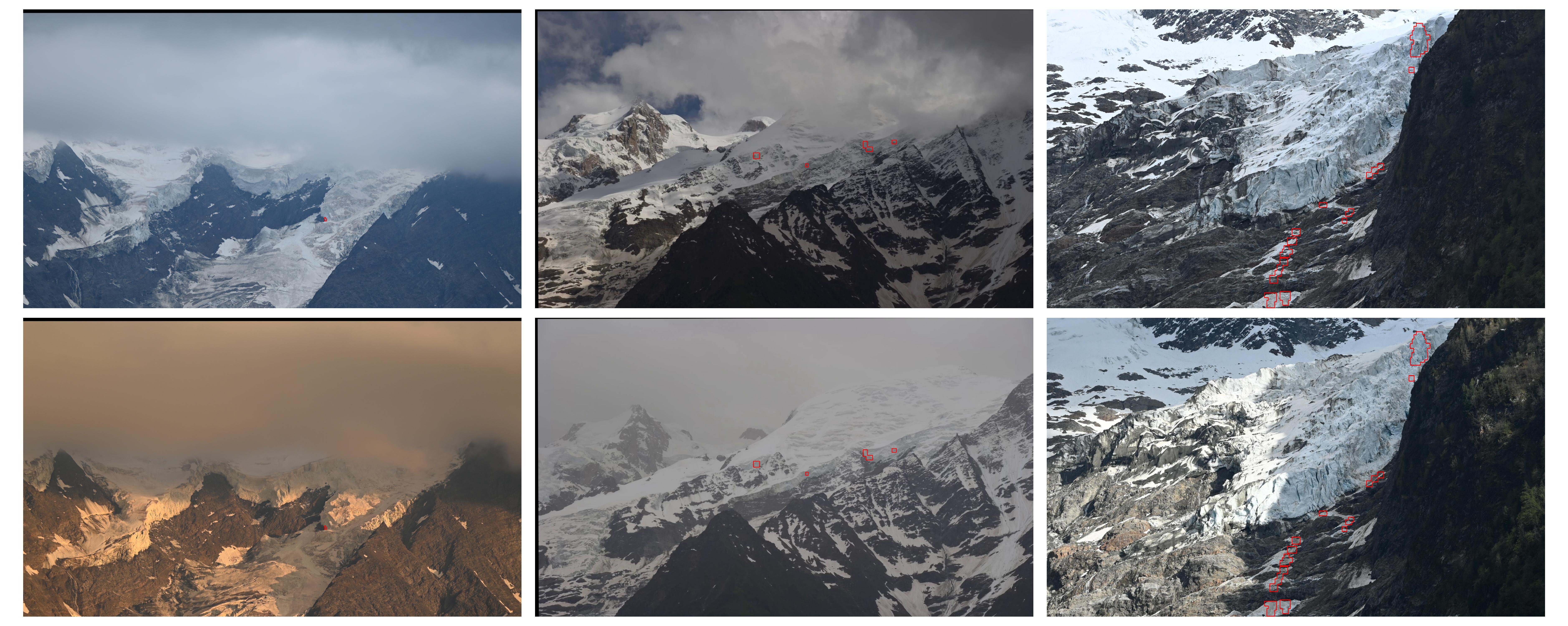}
            \captionsetup{justification=centering}
            \caption{\label{fig:serac_fall}Examples of serac falls (within red boxes) at Taconnaz glacier, involving fog and lighting changes that would make these image pairs unexploitable for classical image correlation approaches.}
        \end{figure*}

        The rapid acceleration of climate change has intensified the occurrence of catastrophic natural hazards, particularly in geologically unstable regions. Relatively small gravitational events (landslides, rock, seracs falls, etc.) have been frequently found to be precursors to such catastrophic hazards~\cite{Precursor1, Precursor2, Precursor3}. Therefore, monitoring them is essential to help us prevent the most dreadful repercussions. A variety of effective sensors have been developed for that purpose, ranging from interferometric radars, lidars, seismometers and extensometers. However, they each face some downsides limiting their scalability such as expensiveness, need of calibration, difficulty of deployment, etc. Consequently, they are primarily deployed on  major risk sites, leaving us with substantial blind spots.
        Meanwhile, terrestrial time-lapse cameras make remote monitoring available at a relatively low cost. Already largely deployed alongside traditional sensors to improve their interpretability through visual context, these cameras are often sufficiently featured to allow high-resolution event detection. Automating their exploitation could further elevate their usefulness, enabling large-scale study of unstable slope dynamics and providing an objective quantification of events.
        Yet, the current state-of-the-art methods for this type of sensor in natural context are based upon image correlation algorithms~\cite{Desrues_2021, image_correlation, image_correlation2}, usually applied on image pairs. Those approaches face significant challenges, the main limitation being their sensitivity to varying weather conditions (such as luminosity change, fog, etc.), thus limiting their usage.
        Relying solely on analogous conditions for both images, like classical image correlation usually does, may signify that we leave out a significant amount of image pairs for the task (Figure~\ref{fig:serac_fall}). More advanced methods in computer vision such as Convolutional Neural Networks (CNN) and more recently Transformer architectures can learn to extract discriminative features invariant to specific conditions. Other advantages of machine learning based approaches are their ability to better cope with unwanted motion or image degradations (e.g., caused by wind, blur, network transmission, etc.) and to require less hyperparameters tuning in operational deployment. These deep learning architectures have proved over the last decade their effectiveness in a wide range of 2D and 3D applications, including monocular depth estimation~\cite{midas_, DepthAnythingv2}, object detection~\cite{YOLO, MaskRCNN} and registration~\cite{LoFTR, RoMa}. 
                
        Thus, change detection, a process that identifies and highlights differences of interest between pairs or sequences of images to produce a change map, increasingly leans on deep learning methods. Recently, this approach has been successfully applied to fields such as remote sensing~\cite{FCNCD, ChangeFormer, BiT, MTP, AnyChange} and autonomous driving~\cite{VL-CMU-CD, ZSSCD, sakurada_2017, DR-TANet}, where real-time analysis of dynamic environments is essential. Yet, its effectiveness relies on a substantial amount of training data. 

        However, in our context, events of interest are scarce, which limits our ability to build a representative annotated dataset. Additionally, annotating such data can be cumbersome due to the requirement for pixel-wise labeling and precise identification of the event's occurrence.\\ Therefore, the objective of this paper is to identify deep learning methods capable of detecting local topology changes in scenes captured by a single ground-based time-lapse camera. In particular, we emphasize the possible strategies to address the scarcity of change events and to enhance generalization across different kinds of events and scenes.
        
\begin{figure*}[ht]
        \centering
        \includegraphics[width = 0.9\textwidth]{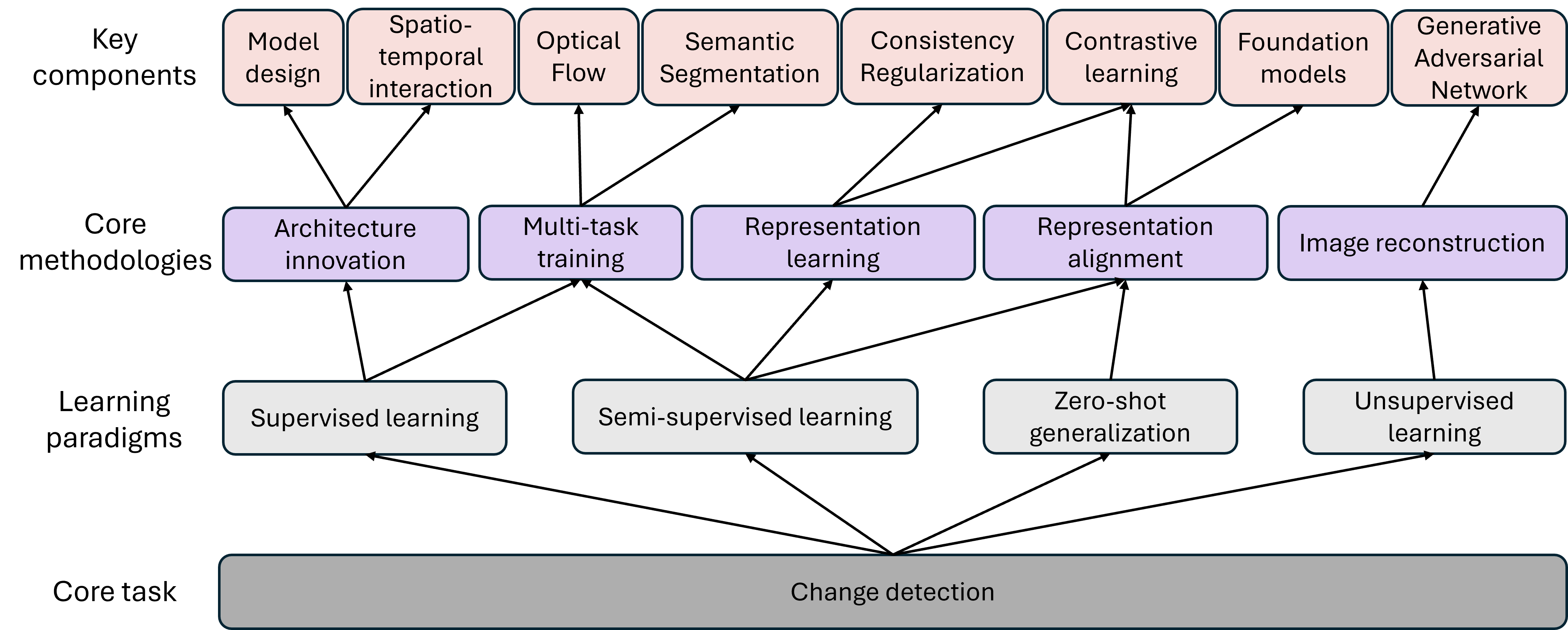}
        \captionsetup{justification=centering}
        \caption{\label{fig:classification_tree}Categorization of change detection methods and their components according to their learning paradigm.}
    \end{figure*}

        The paper is organized as follows: in section~\ref{section:problem_formulation}, we introduce volumetric change detection, a new sub-task for change detection. Section~\ref{section:SOTA} reviews the current state-of-the-art methods in the change detection task (Figure~\ref{fig:classification_tree}), as well as tasks compatible with the stated problem. Section~\ref{section:dataset} depicts the content of a new dataset\textemdash SeracFallDet\footnote{Dataset available at: \url{https://datasets.liris.cnrs.fr/seracfalldet-version1}}\textemdash containing time series of glacier and serac fall annotations. Sections~\ref{section:experiments}~and~\ref{section:results} present the evaluation protocol and model performances of various state-of-the-art methods based on SeracFallDet. Finally, we conclude and discuss the future steps in section~\ref{section:conclusion}.
    
    \section{Problem formulation}\label{section:problem_formulation}
        Given a sequence of ground-level RGB images captured by a single, fixed sensor, our objective is to detect volumetric changes using a pair of images acquired at different times. Those images cover a stable field of view, but the parameters of the sensor are unknown and environmental conditions may vary.
        Volumetric changes are then defined as pixel-wise connected components where the perceived depth varies irreversibly and instantaneously (relative to the frame acquisition rate). This definition includes changes caused by events such as rock falls, serac falls, tree falls, avalanches, etc. It distinguishes them from low-frequency/gradual variations such as ice or snow melt, tree growth and non-structural conditions such as temporary occlusions, not seen as change events in the scope of this study.\\ 
        Mathematically, volumetric changes correspond to:
        \begin{align*}
        \left\{\begin{pmatrix} \tilde{x_{i}} \\ \tilde{y_{i}} \\ 1 \end{pmatrix} = P\begin{pmatrix} x_{i} \\ y_{i} \\ z_{i} \\ 1 \end{pmatrix}, |z_j - z_i| > \epsilon \land |z_k - z_i| > \epsilon \land k - i < \tau\right\}
        \end{align*}
        where:
        \begin{itemize}
            \item ($\tilde{x_{i}}$, $\tilde{y_{i}}$) are the projected coordinates in the image plane at time $i$
            \item ($x_{i}$,$y_{i}$,$z_{i}$) are the 3D world coordinates at time $i$
            \item $P$ is the unknown camera projection matrix
            \item $\epsilon \in \mathbb{R^+}$ is the minimum depth difference threshold to qualify as a change
            \item $i$, $j$, $k \in \mathbb{N}$ are the temporal indices with $i < j < k$
            \item $\tau \in \mathbb{N^{*}}$  is the maximum allowable temporal gap depending on scene dynamics
        \end{itemize} 
    
    \section{Change Detection methods\label{section:SOTA}}
        In this section, we explore how deep learning based approaches tackle key challenges in change detection (Figure~\ref{fig:classification_tree}) and related tasks. Change Detection consists of using pairs or sequences of images captured at different times to extract differences or modifications of interest into a change map through discriminative learned representations. 
        This task is relevant across a broad range of applications in various domains, including remote sensing, environmental monitoring, autonomous driving, and medical imaging. Despite its importance, many applications still rely on manual processes, which are time-consuming and prone to interpretation and/or human error.
        In environmental monitoring, automating change detection enables the study of dynamic phenomena such as slope movements or land-use changes~\cite{FUSU}. By automating this task, continuous monitoring of hazardous regions can help mitigate the impact of catastrophic events, such as landslides~\cite{LandslideMonitoringSatellite}.
        Thus, it has been a topic of interest for several researchers and we will delve into it in the following section.  \\
        
        The first two parts discuss the supervised learning paradigm, with the design of features interaction modules in section~\ref{section:architecture} and multitask learning in section~\ref{section:multitask}. Section~\ref{section:semi-supervised} describes the semi-supervised learning paradigm, mainly based on the consistency regularization framework. Zero-shot learning and unsupervised learning paradigms are then respectively explored in section~\ref{section:zero-shot} and section~\ref{section:unsupervised}. Section~\ref{section:related_task} investigates tasks related to the volumetric change detection, namely dense matching and monocular depth estimation.
        
        \subsection{Supervised learning}
        The supervised change detection task has been extensively explored with the advent of various competitions~\cite{WHU_CD, LEVIR_CD, DynamicEarthNet, VL-CMU-CD, ChangeSim}. Yet, these competitions are in a large part focused on aerial or satellite images, oriented towards building change detection with WHU\_CD~\cite{WHU_CD} and LEVIR\_CD~\cite{LEVIR_CD} datasets, land cover change detection with DynamicEarthNet~\cite{DynamicEarthNet}. Ground-level scene change detection challenges exist, within an urban context with VL-CMU-CD~\cite{VL-CMU-CD} or an indoor context with ChangeSim~\cite{ChangeSim} (Table~\ref{tab:dataset_comparison}). In these fully supervised settings, the goal is to leverage the provided labels as ground truth to learn optimal feature representations and generate accurate change maps.
        
        \begin{table*}[hbtp]
            \centering
            \begin{tabular}{M{0.2\textwidth} M{0.15\textwidth} M{0.18\textwidth} M{0.16\textwidth}}
            \toprule
               Dataset & Point of view / sensor type & \# annotations & Content \\
            \midrule
               DynamicEarthNet~\cite{DynamicEarthNet} & Satellite & \shortstack{1,800\\(54,750 images)} & Various land cover and land use \\ 
            \midrule
               LEVIR\_CD~\cite{LEVIR_CD} & Satellite & 637 & Building change \\
            \midrule
               WHU\_CD~\cite{WHU_CD} & Satellite & \shortstack{1\\(32,507\texttimes15,354 pixels)}  & Building change \\
            \midrule
               VL-CMU-CD~\cite{VL-CMU-CD} & Terrestrial / Vehicle-mounted camera & 1,362 & Urban environment \\
            \midrule
               ChangeSim~\cite{ChangeSim} & Terrestrial / Synthetic & ~20,000 & Industrial indoor environment \\
            \midrule \midrule   
               SeracFallDet (ours) & Terrestrial / Time-lapse camera & \shortstack{1,962\\(6,518 images)}  & Glaciers \\
            \bottomrule
            \end{tabular}
            \captionsetup{justification=centering}
            \caption{\label{tab:dataset_comparison} SeracFallDet comparison with commonly used Change Detection datasets.}
        \end{table*}
        
        \subsubsection{Architecture\label{section:architecture}}
\begin{wrapfigure}[9]{R}{0.5\textwidth}
                    \centering
                    \includegraphics[width = 0.48\textwidth]{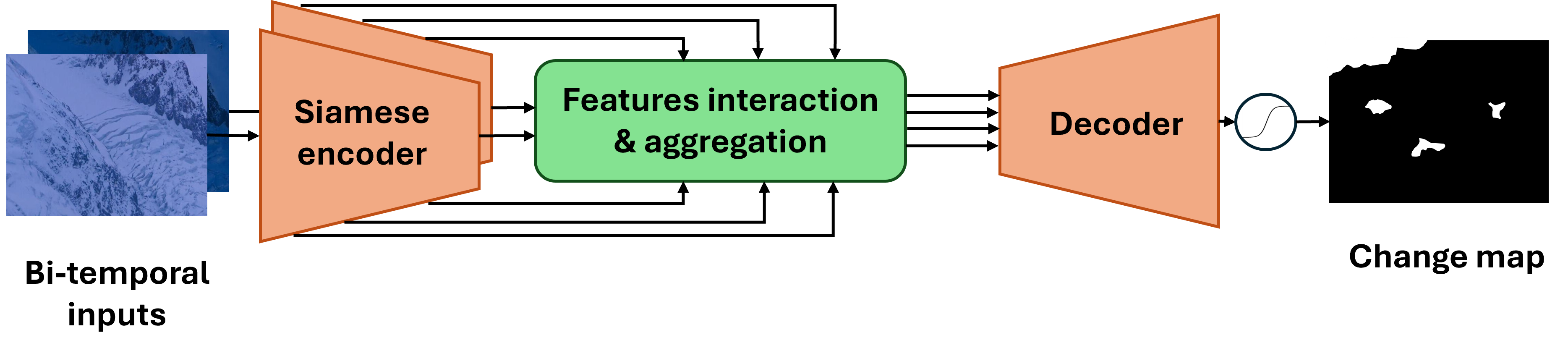}
                    \captionsetup{justification=centering}
                    \caption{\label{fig:CD_model}Standard deep learning architecture for change detection}
            \end{wrapfigure}

            One approach to guide the learning process toward optimal feature representations involves incorporating architectural biases into the model, such as the translation invariance and equivariance properties of CNNs. To this end, recent innovations have focused on designing architectures that ensure feature representations derived from pairs of images are well-aligned, particularly in regions where no changes or irrelevant changes occur~\cite{Changer, DR-TANet, C-3PO, HANet, BiT}. Conversely, in regions where changes of interest do occur, the representations must be misaligned to highlight meaningful differences.
            Nowadays, following~\cite{FCNCD}'s work, most of the deep learning models use similar architectures (Figure~\ref{fig:CD_model}). It consists in using a siamese architecture, where the inputs are processed through the same encoder, potentially as a feature pyramid network (FPN). The model then produces a segmentation map through two stages: the inputs are first down-sampled, then up-sampled back to the original resolution, producing the change detection map.
            In the case of FPNs, the extracted features prior to down-sampling steps are kept, creating multi-scale features. They are then aggregated back at the corresponding resolution during the up-sampling process~\cite{SemiCD, DR-TANet, Changer, HANet, MTP, C-3PO}. Performance gains are mainly obtained by modifying the way the spatiotemporal features interact with specific change detection aggregation modules.
            In that way, FC-Siam-Diff~\cite{FCNCD} uses a difference module (using subtraction) in a fully convolutional U-Net-shaped network~\cite{UNet}, while ChangeFormer~\cite{ChangeFormer} uses a difference module (based on convolution and ReLU activation function of concatenated features) at each stage of its FPN. \\
            C-3PO~\cite{C-3PO} considers the change detection task as a sub-task of semantic segmentation by solely inserting a spatiotemporal aggregation module. To this end, they designed two modules placed between the encoder and the decoder of a segmentation model. The first module, \textit{Merge Temporal Features}, reveals object appearance, disappearance and exchange through feature subtraction depending on the temporal order of the inputs. The second module, \textit{Merge Spatial Features}, aggregates the multi-scale features through summation and up-sampling.
    
            With the relatively recent breakthrough of Attention modules~\cite{Attention} and Vision Transformers~\cite{ViT}, which ChangeFormer~\cite{ChangeFormer} and BiT~\cite{BiT} are using, attention-based aggregation modules have been developed~\cite{DR-TANet, HANet}. To that end, DR-TANet~\cite{DR-TANet} uses a kernel-sized (local) cross-attention module with the bi-temporal feature maps as an aggregation module, using a larger kernel size to produce finer feature maps. HANet~\cite{HANet}, on the other hand, concatenates the bi-temporal features and instead processes them through dilated convolutions to increase the receptive fields. The concatenated outputs are then enhanced with global self-attention and row/column-attention. \\
            Sophisticated interactions have been further explored with Changer~\cite{Changer} with the design of multiple interaction and fusion layers. The interaction layers aim to enhance the features produced by the FPN, by exchanging channels (channel exchange), spatial features (spatial exchange), or channel attention from the summed features. The fusion layer involves the final network’s feature interaction, estimating the optical flow to warp pixels, potentially allowing small misregistration, before subtraction.
            
            To summarize, three fundamental components are commonly highlighted in the design of modern architecture.\\
            The first component is the siamese architecture, which has become a consensus in recent change detection models, particularly for single-sensor inputs. This design enables models to learn photometric invariance and improve efficiency, with CNNs being used to extract local patterns and/or Transformer blocks to expand the receptive field.\\
            Similarly consensual, the second component is the FPN, primarily used to detect changes at various scales.\\
            The third main component is the interaction module, which is responsible for incorporating temporality into the architecture, and remains a research topic. Some approaches specifically include explicit bias through feature subtraction~\cite{FCNCD, C-3PO, BiT}, imposing strict constraints compared to more flexible parameterized modules making use of concatenation and convolutional block~\cite{ChangeFormer} or attention mechanism~\cite{DR-TANet, HANet}.\\
            All these architectural improvements have allowed networks to continually improve, achieving excellent performances on datasets such as building CD from satellite imagery~\cite{WHU_CD, LEVIR_CD} and urban change detection~\cite{VL-CMU-CD, Sakurada_2020}. However, these datasets are far from terrestrial natural hazard images, which hinders their transfer capacities without extensive data.
            
        \subsubsection{Multi-task learning\label{section:multitask}}
            Change detectors excel with sufficient supervision, but they often struggle with data scarcity, leading to poor generalization. While multi-task learning can help solve some task-specific challenges, it can also leverage additional auxiliary tasks and their corresponding datasets alongside the primary task, extending same or similar domain datasets. This approach involves jointly training a model on multiple labeled tasks, potentially on several datasets. The learned features are then constrained to extract as much information as possible to satisfy the different tasks' objectives. Introducing a sub-task guides the feature extraction in a direction that aligns with both the main and auxiliary objectives. After the training process, the auxiliary task head can either be discarded~\cite{MTP} or, more commonly, retained to enhance performance on the main task head~\cite{sakurada_2017, Sakurada_2020, SimSaC}. However, it can also be detrimental to the model's performance when performed on incompatible tasks, guiding the model’s features in conflicting directions~\cite{MultiTaskOverview}. \\
            Following this paradigm, MTP~\cite{MTP} uses a foundation model architecture (InternImage-XL~\cite{InternImage} and ViT~\cite{ViT} + RVSA~\cite{RVSA}) and compares its performance using multi-task pretraining on SAMRS dataset~\cite{SAMRS} containing rotated objects, instance segmentation and semantic segmentation annotations in remote sensing images. Their goal is to provide a remote sensing foundation model, which can be used on downstream tasks once it has been fine-tuned. For each task, the encoder is shared but a distinct decoder head is used. The multi-task pretraining demonstrates slight improvements compared to fine-tuning solely on datasets with few training iterations (otherwise, the impact of pretraining tends to diminish).
            
            On the other hand, some researchers have sought complementary yet distinct tasks\textemdash such as optical flow\textemdash to improve change detection performance. As in the previous work~\cite{sakurada_2017}, SimSaC~\cite{SimSaC} proposes to combine motion estimation through optical flow with change detection training to boost performance. Since such multi-label datasets do not exist, they create a synthetic optical flow dataset using a cut-and-paste method alongside data augmentation. Using local correlation and an optical flow decoder (based on FlowNet~\cite{FlowNet}), they estimate the optical flow, and use it both as a pixel warper to account for slight misregistration and as an additional input to the change detection decoder. Thanks to the integration of both tasks, they significantly improve performance, particularly on the more challenging portions of the test dataset.
            
            To summarize, the multi-task learning framework has proven highly effective in improving change detection performances by handling task-specific challenges. Studies have shown that data from the same domain provided by auxiliary tasks, through the use of additional task-specific heads, enables models to extract information meaningful for the main task. This particularly applies in scenarios with limited data~\cite{MTP}. However, its impact is limited when sufficient labeled data is already available. Nevertheless, benefiting from a substantial and fully labeled dataset for ground-level applications in natural environments might not be affordable due to the extensive annotation process required.
        
         \subsection{Semi-supervised learning\label{section:semi-supervised}}
        To face the shortage of annotation, semi-supervised methods have been explored, focusing on the learning framework to achieve near-supervised performance. The main idea is that in contexts where labeled annotations are scarce, unlabeled data still contains valuable information that can enhance the model’s representational capabilities. Therefore, the goal is to design tailored loss functions (e.g., consistency regularization~\cite{SemiCD, SemiCD-VL, UniMatch, UniMatchv2} and task-alignment~\cite{SimSaCv2}) that maximize the usefulness of unlabeled data. \\   
        As a continuation of their work~\cite{SimSaC, SimSaCv2}, SimSaC authors aim at maintaining as much performance as possible using their synthetic dataset and the target dataset with little to no annotation. They use a teacher-student framework with a shared encoder, where the teacher and student both learn from synthetic data with a supervised loss. However, the teacher additionally learns from the target dataset through unsupervised feature alignment (between the optical flow and change map), as well as change regularization and edge regularization. Change regularization enforces high similarity in the no-change regions, while edge regularization ensures that the boundaries of change areas coincide with those in the original images. Meanwhile, the student also learns from the teacher's distillation on the target domain, allowing modifications to its predictions and handling the domain gap with synthetic data.
        Complementary to works on multi-task learning, the main framework explored in semi-supervised settings focuses on consistency regularization (CR). The CR hypothesis relies on both the cluster assumption, which states that data in high-density regions belong to the same class, and the low-density separation assumption, which states that the decision boundary between high-density regions is located in low-density regions~\cite{SemiCD}. In other words, regions where classes are mixed are not dense. Thus, it is possible to reinforce the decision boundaries from unlabeled data where the model must produce the same (or similar) output while the input is perturbed.
        SemiCD authors~\cite{SemiCD} have pioneered the application of this technique to change detection by injecting small perturbations (uniform noise, channel masking, spatial masking, etc.) into the hidden features space and using a reconstruction loss between the perturbation-free and weakly perturbed outputs. This unsupervised loss further helped the learning process, thus requiring less labeled data to achieve near fully supervised performance.
        Following this implementation, UniMatch~\cite{UniMatch} and later UniMatchv2~\cite{UniMatchv2} further explored the CR on semantic segmentation (which can be considered a particular case of change detection). Their approach introduces three parallel streams processing the same weakly perturbed input:
        \begin{itemize}
            \item One stream applies feature-level perturbations (e.g., dropout or uniform noise)
            \item Two streams undergo strong pixel-level perturbations (e.g., color jittering)
        \end{itemize}\hfill\break\indent
        The model is then trained by using a cross-entropy loss between the pseudo-label (derived from the weakly perturbed input) and the outputs of the three streams. To reduce noise, only pseudo-labels with the highest confidence are retained as unsupervised loss. \hfill \break \indent
        The two strongly perturbed streams exhibit high similarity with contrastive learning. In fact, contrastive learning is a self-supervised approach that aims at bringing closer the feature representations of similar samples, and pushing apart those of dissimilar samples. In this case, since the dual-stream features are supervised using a shared pseudo-label for two distinct views of the same input, these features should be similar, aligned with the core objective of contrastive learning.
        In their follow-up work, UniMatchv2~\cite{UniMatchv2} increased their performance by using a modern backbone architecture, namely DINOv2~\cite{DINOv2}, instead of ResNet~\cite{ResNet}. Taking advantage of the robustness of DINOv2, the authors applied the feature perturbation directly to the dual-stream branches with complementary dropout, randomly masking out half of the features in both streams. This approach encourages the model to learn more robust features and fully leverage the feature space. A second update involves using an exponential moving average of the student’s weights to update the teacher model, producing higher-quality and more stable pseudo-labels.
        Meanwhile, SemiCD-VL~\cite{SemiCD-VL} combines the CR framework (similar to UniMatch~\cite{UniMatch}) with the zero-shot capabilities of a Visual Language Model (VLM) to produce pseudo-labels. The VLM produces segmentation maps using a predefined vocabulary, split into foreground and background sets. VLM pseudo-labels are then derived from instances where objects appear or disappear between the two sets. The final loss in SemiCD-VL integrates CR losses, VLM-based pseudo-labels, and a contrastive loss to improve performance. \hfill \break \indent
        To summarize, the consistency regularization has proven to be a powerful framework for exploiting unlabeled data. By leveraging weak and strong augmentations, it enables models to learn invariance while accurately detecting changes from a limited number of annotations. Exploring meaningful perturbation, whether in pixel space~\cite{SemiCD, UniMatch, UniMatchv2} or feature space~\cite{UniMatch, UniMatchv2}, in a complementary manner, has led to more robust models. Notably, models must effectively utilize their feature dimensionality to encode information correctly and avoid over-relying on a limited subset of dimensions to highlight change regions (e.g., primary objective of feature dropout~\cite{UniMatch, UniMatchv2}). Meanwhile, the teacher-student framework has been used for two key purposes: domain adaptation~\cite{SimSaCv2} and stable pseudo-label generation~\cite{UniMatchv2}. Further task-related cues, such as the edge regularization~\cite{SimSaCv2}, also help to enhance the extraction of meaningful information from unlabeled data.
        The use of unsupervised loss in this context has shown significant benefits in helping models to learn key task-related information from a small amount of labeled data, leading to better generalization (cross-dataset zero-shot or few-shot transfer)~\cite{SemiCD, SemiCD-VL}. However, the effectiveness of this approach relies on the assumption that the unlabeled data shares a similar distribution with the labeled data to ensure meaningful information extraction. In our context, while a significant amount of unlabeled images without changes are available, the labeled set primarily consists of pairs with changes. This distribution mismatch may bias the generated outputs during inference, leading to a tendency to predict fewer changes.
         
        \subsection{Zero-shot generalization\label{section:zero-shot}}
        Another way to address data scarcity is to leverage models with heavy pretraining on distinct tasks or large datasets, in order to enhance generalization and enable 'zero-shot' predictions. Here, our definition of 'zero-shot' refers to the use of a pretrained model in an unseen task or domain without fine-tuning. Thus, the key challenge lies in bridging the gap between the training and target task/domain and mostly involves the use of foundation models.
        Following that principle, Any Change~\cite{AnyChange} attempts to leverage SAM and SAM2~\cite{SAM, SAM2}, which are trained to produce class-agnostic segmentation masks, for zero-shot prediction in the change detection task without fine-tuning. Since change detection involves temporality, they proposed a bi-temporal latent matching approach to bridge the gap between the two tasks. It involves computing the masks for images taken at two different times. After extracting the embedding spaces of both images and up-sampling them to the original image resolution, they compute the average spatial features of the masks in the embedding space. If the cosine similarity between the average mask features is below a pre-defined threshold, they are considered change masks.
        
        GeSCF~\cite{GeSCD} uses bi-temporal SAM embedding at intermediate layers to compute embedding similarities and generate pseudo-masks from dissimilarities. These pseudo-masks are generated using an adaptive threshold that takes into account the skewness of the similarity distribution. Their geometric matching method then computes the intersection between SAM’s generated masks and the pseudo-masks to obtain object-wise changes. Finally, the change map is refined by extracting semantic information from the bi-temporal mask embeddings using cosine similarity, further discriminating pseudo-mask noise.
        
        In a similar fashion, Zero-Shot Scene Change Detection~\cite{ZSSCD} considers change detection as a tracking task. Using SAM~\cite{SAM} as s segmentation model and DEVA~\cite{TrackAnything} for mask tracking, they avoid supervised learning. The goal is for DEVA to track the masks produced by SAM in the forward timeline to detect masks' disappearance, and in the reversed timeline to detect newly appeared masks, both of which are defined as changes.

        ~\cite{ZeroSCD} similarly uses SAM to generate multiple masks on images from both time points, as well as PlaceFormer~\cite{PlaceFormer}, a visual place recognition (VPR) model, to extract the images’ homography. Additionally, since VPR models are trained to be invariant to illumination changes, they use their extracted features to produce a change heat map based on Euclidean distance to further highlight dissimilar features in regions where changes have occurred. Then, by comparing the overlap between the generated masks and the heat-map, the mask is considered to indicate a meaningful change.
        
        With these different works, pretrained models' latent spaces have been extensively probed to find cues and exploit them in the change detection task (embedding similarities~\cite{AnyChange, GeSCD, ZeroSCD}) or integrated into a pipeline as a block~\cite{ZeroSCD, PlaceFormer}.
        The downside of these methods is that their performance depends heavily on the zero-shot capabilities of foundation models (e.g., SAM, DINOv2, etc.), which may not hold when facing datasets containing entirely different types of images, such as natural scenes. Another typical issue is that those models may learn invariance to certain properties useful in the Change-Detection task.\hfill \break \indent 
        Compared to other zero-shot methods, the VPR model in~\cite{ZeroSCD} was evaluated on a dataset similar to its training data. This similarity may limit its generalization to different types of images, potentially hindering performance in real-world applications.
        
    \subsection{Unsupervised learning\label{section:unsupervised}}
        Another research axis involves alleviating the challenges of unsupervised learning by exploiting unlabeled data and task-related cues to detect changes. This approach has been mostly explored by using synthetic labels or detecting outliers. The authors of~\cite{Image_few_samples} propose to perform it by handcrafting augmentations to simulate changes, although the results often lack realism. In contrast,~\cite{Self-pair} attempts to produce more realistic synthetic changes by copying and pasting regions with Fourier transform-based blending.

        Similar to Anomaly Detection,~\cite{Image_reconstruction} uses a reconstruction strategy to detect changes, by expecting a high reconstruction error where a change is located. In addition, a generative adversarial network (GAN) and photometric transformations are used to make the model invariant to style changes.
        However, since these datasets or transformations are still unrealistic, the main issue lies in the domain shift when applied to real images. This must be addressed through various domain adaptation strategies designed to bridge the gap with the target distribution.
        S2-cGAN~\cite{S2-cGAN} uses a conditional GAN differently: from the initial patch and added perturbation noise, the authors generate a synthetic patch, and use a discriminator to compare it to the initial patch. The assumption behind this strategy is that both generated and changed pixels will be treated as outliers at inference time, allowing the discriminator to identify them accordingly.
        
        \subsection{Related tasks\label{section:related_task}}
        As a complement to the tasks mentioned in section.~\ref{section:multitask}, or in order to help in the zero-shot capabilities discussed in section.~\ref{section:zero-shot}, some related methods linked to volumetric variations may assist the change detection model in this task. In fact, the zero-shot performances of monocular depth estimation and dense matching models are remarkable, potentially offering valuable insight in other contexts.
        \subsubsection{Dense and Semi-Dense Matching}
            The dense matching task consists of associating pixel tiles from one image with corresponding tiles in a second image, despite significant viewpoint and potential illumination changes. Since stereo matching reconstructs depth from pairs of similar images with 3D transformation (such as rotation and/or translation), dense matching datasets provide implicit 3D information that the model can learn from. This task is characterized by the search for correspondences between images. Such correspondences may be absent due to volume loss or occlusions, resulting in unmatched pixel pairs. However, in change detection, the cameras are assumed to be static, even though slight misregistration or subtle object motion may still occur, leading to fewer constraints for the feature extraction.
            The main state-of-the-art algorithm derives from the LoFTR~\cite{LoFTR} architecture, where the authors created a keypoint detector-free dense matching model by extracting features with a CNN at both coarse and fine levels. Using the capabilities and receptive field of transformers (which include self-attention and cross-attention), they enhance the coarse features to facilitate matching based on feature similarity. The matched patches are then locally processed through a transformer block using fine features to find the precise pixel in the second image corresponding to the patch center in the first image.
            
            TopicFM and TopicFM+~\cite{TopicFM, TopicFM+} are two of the models based on the LoFTR architecture, where the authors added a learnable topic discriminator module. The assumption is that knowing the semantics (topics) of an image helps the model to match pixels, because it is unlikely for pixels with unrelated semantics to match. The topic module consists in finding the topic correspondence of each patch and sampling the topic distribution (based on a multinomial distribution), in order to identify co-visible topics and ultimately enhance the matchability of these co-visible topics’ features. In the subsequent work~\cite{TopicFM+}, the features are instead merged with the topic embedding for feature enhancement, and the fine-level refinement from the LoFTR architecture is performed through an MLP-Mixer network~\cite{MLP-Mixer}.
            
            ASpanFormer \cite{ASpanFormer} is also based on the LoFTR architecture and introduces optical flow estimation as a Gaussian distribution, along with a Global-Local Attention (GLA) module between the encoder and the coarse matching module. The idea is to use the optical flow variance within a patch as an indicator of the difficulty of coarse matching, allowing the use of an adaptive patch size for local attention. The coarse features are enhanced through GLA (instead of the LoFTR transformer) by further down-sampling the coarse features into two scales. The coarsest scale is processed with global cross-attention, while the medium and initial scales are processed through local cross-attention with a spatially adaptive window size depending on the estimated flow variance.
            
            Varying from the LoFTR architecture, DKM~\cite{DKM} considers instead dense feature matching as a regression task. Using an FPN and the posterior mean of a Gaussian process, they estimate the warp of the embedding pixels at coarse scales, which is then decoded with CNNs to predict both the pixels warp and their corresponding match confidence. The predictions are subsequently refined through a second stage by a CNN that estimates the displacement needed to achieve accurate alignment at finer scales. Thanks to this formulation, they provide true pixel-wise dense matching while preserving match confidence.
            
            In their following work, RoMa~\cite{RoMa}, the authors focused on making the models more robust with the help of the more modern DINOv2 backbone~\cite{DINOv2}, a transformer architecture to decode the coordinate embedding, and a distinct CNN to extract fine features (refining the coarse predictions). On top of that, they integrate robust loss optimization for both coarse and fine scales to train the dense matcher.
            
            In this field, the primary scientific challenges revolve around the efficiency~\cite{TopicFM, TopicFM+} and the reliability~\cite{ASpanFormer, RoMa, DKM} in the keypoints matching process.
            
        \subsubsection{Depth estimation}
\begin{wrapfigure}[19]{L}{0.5\textwidth}
                \centering
                \includegraphics[width = 0.48\textwidth]{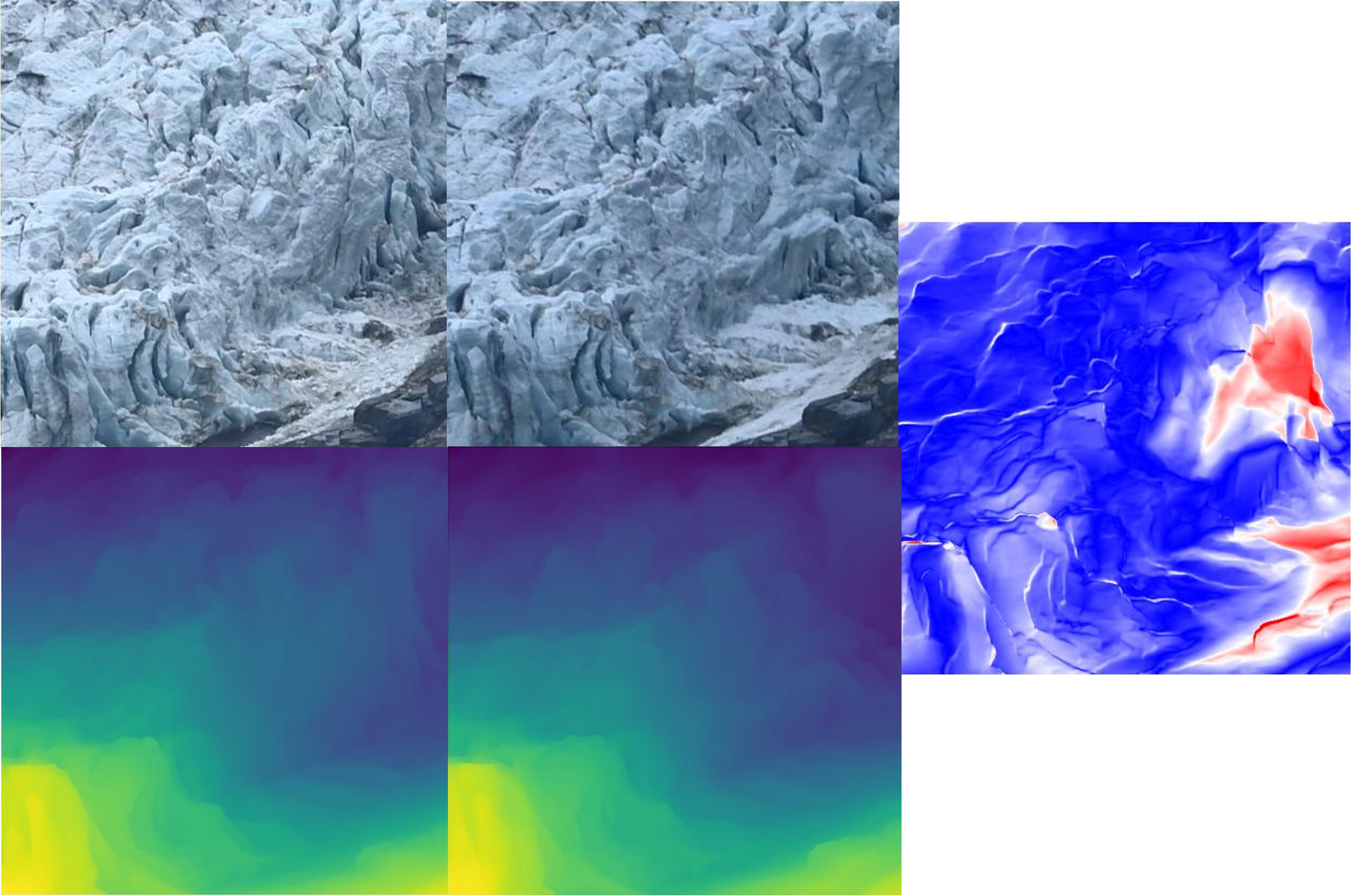}
                \caption{Illustration of DepthAnythingv2~\cite{DepthAnythingv2} inferences on two image patches (top), their corresponding estimated depth (bottom) and the absolute depth difference. (right)\label{fig:depth_diff}}
            \end{wrapfigure}

            Contrary to dense matching, monocular depth estimation (MDE) methods learn explicitly 3D information by exploiting a single image to infer depth. Provided MDE is consistent enough between image pairs and sufficiently resoluted, this class of methods might reveal important 3D information by highlighting the differences between the two produced depth maps (Figure \ref{fig:depth_diff}). 
            
            MiDaS~\cite{midas_} aims at producing a depth foundation model with zero-shot cross-dataset transfer capabilities by training on multiple datasets, thereby achieving performance closer to real-world conditions by mitigating each dataset's inherent bias. However, challenges arise since depth values are not on the same scale (due to varying extraction methods) and use different depth representations (e.g., direct vs. inverse). To address this, the authors designed loss functions to handle these inconsistencies, including a scale- and shift-invariant L1 loss and a strategy that trims the 20\% of depth values with the highest residuals between prediction and ground truth. Additionally, they introduced an edge regularization loss to encourage alignment between edges in the ground truth and those in the predicted depth map. Since the datasets vary in size, it is unclear whether they should be naively combined during training. To explore this, the authors experimented with mixing the datasets evenly and applied Pareto-optimal multi-task learning which, appeared to improve performance compared to the naive approach.
            
            Subsequently, DepthAnything~\cite{DepthAnything} aims at improving upon MiDaS~\cite{midas_} by employing a teacher–student framework, where the teacher is trained following the MiDaS methodology. However, only five datasets were selected for training and were processed using semantic segmentation to assign a disparity value of 0 to the sky (the furthest depth). The teacher then produces pseudo-labels for the student model, which is initialized with new weights to avoid inheriting the teacher’s biases. The student's performance was improved through strong input perturbations, thus challenging the student model to reconstruct the teacher’s predictions. Additionally, to preserve semantic information, the authors distilled features extracted from DINOv2 using cosine similarity, retaining only those with similarity below a predefined threshold, acknowledging that 3D semantics do not perfectly align with 2D semantics.
            
            In their subsequent paper DepthAnythingv2~\cite{DepthAnythingv2}, they identified that the first version lacked fine-grained detail, primarily due to the limitations of the training data. To address this, they constructed a synthetic dataset using computer graphics, providing precise 3D ground truth, which was used to train the teacher model and generate pseudo-labels on real-world datasets, thus partially mitigating domain shift. To further reduce this domain gap, the model ignores 10\% of regions with the highest L1 reconstruction error. Additional improvements were made by incorporating MiDaS’ edge regularization loss and applying feature alignment using DINOv2 embeddings.
            
            For the task of MDE, a significant number of datasets exist, and combining them to effectively mitigate their biases made generalization to previously unseen domains possible~\cite{midas_}. The use of synthetic generation with foundation models and a teacher-student approach~\cite{DepthAnythingv2} further improved the generalization capabilities and fine-grained details seen in images. \\

    \subsection{State-of-the-art discussion}

   In conclusion, the change detection task has been a topic largely investigated in various fields such as autonomous driving and remote sensing, leading to significant competition and numerous advances in learning strategies. The works in supervised learning helped to find optimal feature interactions, incorporating spatiotemporal understanding in the change detection model's behavior. This interaction has been further complemented with multi-task learning and allowed the usage of valuable information from other datasets. Notably, SimSaC~\cite{SimSaC} incorporates optical flow as an auxiliary task, useful when handling slight misregistration, which is common in the change detection task. The semi-supervised framework and the integration of consistency regularization~\cite{SemiCD, UniMatchv2, SemiCD-VL} achieve remarkable performances when facing annotation scarcity. Recently, zero-shot change detection has been made available by leveraging the capabilities of SAM~\cite{SAM, SAM2} by extensively using its latent space~\cite{AnyChange, GeSCD}.
   However, to our knowledge, this task has been subject to little exploration within domains where it could be valuable, such as natural hazard monitoring or medical imaging. This creates a substantial domain gap that even foundation models struggle to bridge. The following sections aim at bridging this gap through the creation of a dataset and a benchmark protocol dedicated to natural hazard monitoring.

    \section{Dataset\label{section:dataset}}
        In this section, we present the new dataset SeracFallDet\footnote{Dataset available at: \url{https://datasets.liris.cnrs.fr/seracfalldet-version1}}, a dataset with time series of glaciers and serac fall labels, to illustrate the stated problem on volumetric change detection and assess the domain gap with the current datasets. \\
    To our knowledge, there is currently no dataset containing labeled natural geomorphological changes from a terrestrial point of view. Therefore, we created a manually annotated dataset of natural hazard events.
    Each sequence has been registered using a variation of the pipeline proposed in~\cite{Lemaire_2021}, and each event between two images has been annotated using polygonal boxes (rather than pixel-wise labels). This results in 1962 pairs of high-resolution images from 11 different sites. Each site may contain a different timeline from multiple years and is mostly located in the French Alps (Table \ref{tab:dataset_content}). Unlike common change detection datasets such as~\cite{LEVIR_CD, WHU_CD, VL-CMU-CD, ChangeSim}, our sequences span longer periods, allowing for the extraction of meaningful temporal patterns as in~\cite{DynamicEarthNet, FUSU}, though with fewer distinct scenes (see Table~\ref{tab:dataset_comparison}). This dataset also distinguishes itself from others in that the events of interest directly influence an object's local topology rather than its entire structure. Even though this dataset is specific to serac fall, we argue that methods developed in this context hold significant potential for broader applications with similar structural changes, such as landslides. 
    \begin{figure*}[htbp]
        \centering
        \includegraphics[width = 1.0\textwidth]{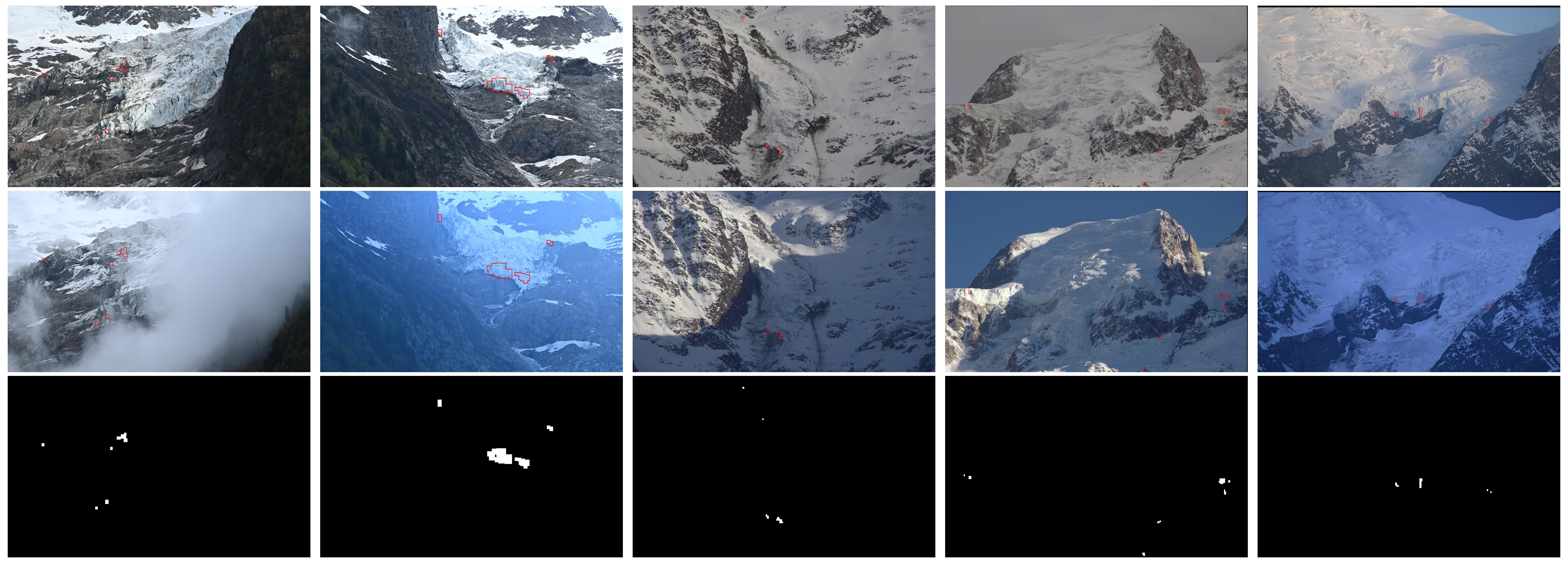}
        \captionsetup{justification = centering}
        \caption{\label{fig:supplementals} Example of image pairs under different weather conditions (first two rows) in different scenes with their corresponding changes (in red and third row).}
    \end{figure*}

    \begin{table*}[ht]
        \centering
        \begin{tabular}[]{ M{0.15\textwidth} M{0.05\textwidth} M{0.12\textwidth} M{0.12\textwidth} M{0.11 \textwidth} M{0.08 \textwidth} }
        \toprule
            Sites& Year & \# images& \# annotations& Image resolution & Split\\ \midrule
            Taconnaz terminus & 2024 & 826 & 174 & 3850\texttimes1900 & Test \\ \midrule
            Bossons terminus 1 & 2024 & 827 & 129 & 2750\texttimes1500 & Test \\ \midrule
            Bossons terminus 2 & 2024 & 340 & 50 & 3840\texttimes2160 & Test \\ \midrule
            Glacier rond & 2024 & 340 & 19 & 3840\texttimes2160 & Test \\ \midrule
            Mont Blanc du Tacul & 2024 & 340 & 53 & 3840\texttimes2160 & Test \\ \midrule \midrule
            Grands Mulets terminus & 2024 & 822 & 140 & 2500\texttimes1200 & Validation \\ \midrule \midrule
            Taconnaz 1 & 2024 & 333 & 74 & 3840\texttimes2160 & Train \\ \midrule
            La Griaz glacier & 2024 & 123 & 125 & 3840\texttimes2160 & Train \\ \midrule
            Pyramid plateau & 2024 & 225 & 112 & 3840\texttimes2160 & Train \\ \midrule
            Mont Blanc & 2024 & 1230 & 769 & 2300\texttimes1460 & Train \\ \midrule
            Taconnaz 2 & 2025 & 1112 & 317 & 4000\texttimes2400 & Train \\ 
        \bottomrule
        \end{tabular}
        \captionsetup{justification=centering}
        \caption{\label{tab:dataset_content}SeracFallDet dataset content description.}
    \end{table*}
    
\begin{wrapfigure}[19]{R}{0.5\textwidth}
            \centering
            \includegraphics[width = 0.47\textwidth]{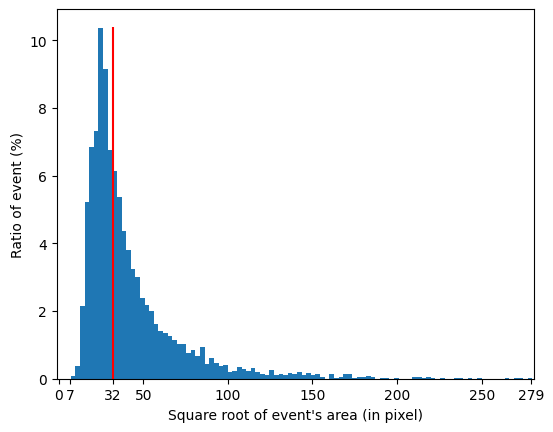}
            \captionsetup{justification=centering}
            \caption{\label{fig:event_size} Distribution of event sizes in SeracFallDet dataset. Median is displayed in red.}
    \end{wrapfigure}

   Since the SeracFallDet dataset involves natural images from a terrestrial viewpoint, it poses specific challenges for computer vision models. In fact, as far as we know, no dataset includes enough natural images to pretrain a foundation model, which makes them suffer from a domain gap. On the other hand, the main challenges with this dataset include varying weather conditions (Figure~\ref{fig:supplementals}), such as extreme lighting and shadow changes, fog, and significant seasonal variation; the small sizes of events (Figure~\ref{fig:event_size}); and the presence of noisy labels. Such factors remain difficult to handle, especially given the limited number of training pairs.
    
    The SeracFallDet dataset is decomposed into training, validation, and test sets. The images of each monitoring site belongs exclusively to a single set to avoid biasing the evaluation procedure towards an already seen set of images and favor model generalization. The assignment of monitoring sites to different sets of SeracFallDet is indicated in Table \ref{tab:dataset_content}.
    
    \section{Experiments\label{section:experiments}}
        In this section, we first define the evaluation metrics used to assess the performance of the models. Following this, we introduce and justify the selection of various state-of-the-art methods that will be evaluated on the SeracFallDet dataset in the subsequent section. This includes a discussion on how these methods, originally designed for broader tasks or applications, can be repurposed to better address the specific demand of the SeracFallDet dataset and, more generally, tasks related to monitoring geomorphological change from ground-based imagery.
        
        \subsection{Evaluation metrics}

\begin{figure}[ht]
        \centering
        \includegraphics[width = 0.47\textwidth]{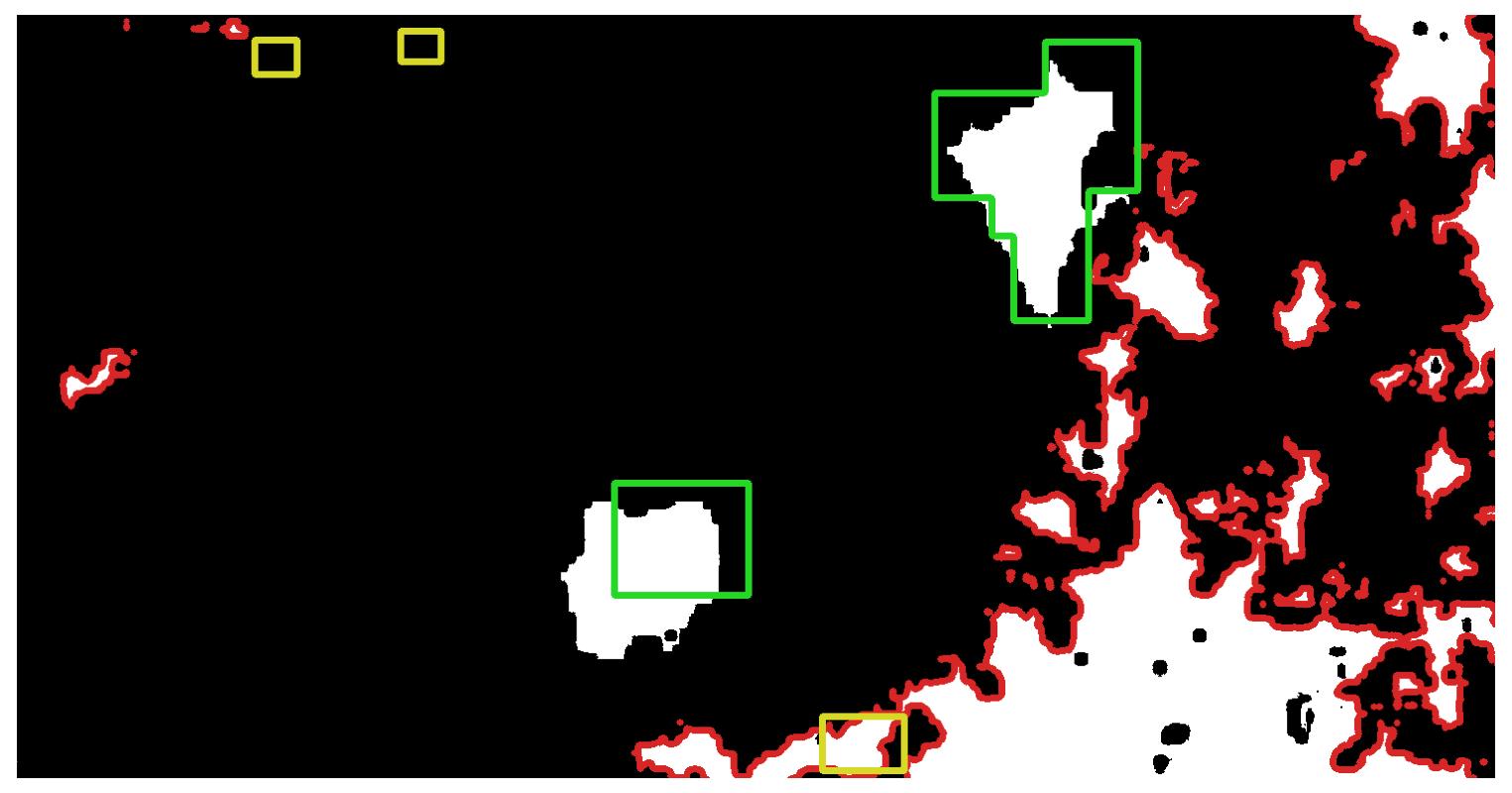}
        \captionsetup{justification=centering}
        \caption{\label{fig:IoU} Illustration of true positive (in green), false positive (in red) and false negative (in yellow) in change detection with IoU threshold above 25\%.}
    \end{figure}

    To correctly highlight the performances of various models, the metrics must be well adapted to the objective the models are trained for. In this context, the core metric in the change detection task is the pixel-wise F1-Score which represents a balance (as harmonic mean) between the precision (indicating correct predictions) and the recall (indicating missing predictions). Another common metric is the binary intersect over union (IoU), slightly penalizing more underestimations (false negatives) and overestimations (false positives). However, in the context of natural hazard monitoring, recall and precision should not be equally weighted. For example, missing a hazardous event (low recall) could have more severe consequences than misclassifying a non-event (low precision). In this case, both metrics provide valuable information about the model's behavior. However, due to the polygonal nature of the annotations, the counts of true positives may be slightly overestimated, while false negatives and true negatives may be underestimated\textemdash particularly in regions near label boundaries\textemdash when compared to noiseless ground truth. \\
    While pixel-wise metrics like F1-Score and IoU are useful, they may not fully capture the operational requirements of natural hazard monitoring, where the localization and identification of discrete events are often more critical than pixel-level accuracy. \\
    To address these limitations, we propose to use an event IoU metric, similar to those used in object detection, which evaluates the overlap between predicted and ground-truth regions as individual events and is computed as follows. Using the change map prediction and ground truth polygons, each connected component is considered an individual event. All predicted events that spatially intersect with a ground-truth polygon are merged into a single prediction mask. The event IoU between this merged prediction and the corresponding ground-truth polygon is then calculated. If the IoU exceeds a predefined threshold (e.g., 0.25), the prediction is classified as a true positive; otherwise, it is counted as both a false negative and a false positive. Thus, ground-truth polygons with no intersecting predictions are counted as false negatives, and predictions without any ground-truth intersection are counted as false positives (Figure~\ref{fig:IoU}). \\
    Although ground-truth annotations may contain some noise, they are assumed to be sufficiently accurate for reliable event-level evaluation. This metric offers a more robust assessment by reducing the influence of annotation or prediction noise (Section~\ref{section:dataset}) and by ensuring that predicted regions correspond to well-localized changes rather than coincidental pixel overlaps. \\
    
    \subsection{Baseline methods\label{section:baseline_methods}}
    The evaluation focuses on models designed to operate effectively under limited annotated data or demonstrate strong generalization capabilities. We considered four categories:
    \begin{itemize}
      \item Semi-supervised methods (SemiCD~\cite{SemiCD} and UniMatchv2~\cite{UniMatchv2}), which leverage both labeled and unlabeled data and do not require a foreground-background separation~\cite{SemiCD-VL}.
      \item Zero-shot change detectors (AnyChange~\cite{AnyChange} and GeSCF~\cite{GeSCD}), which are specifically developed for inference without fine-tuning.
      \item Task-related methods (RoMa~\cite{RoMa}, LoFTR~\cite{LoFTR}, ASpanFormer~\cite{ASpanFormer}, TopicFM+~\cite{TopicFM+}, DepthAnythingv2~\cite{DepthAnythingv2}) adapted to address this change detection objective.
      \item Supervised change detector (SimSaC~\cite{SimSaC}, C-3PO~\cite{C-3PO}) used for reference; C-3PO for its simplicity and SimSaC for handling motion and deformation, typical in glacier environments.
    \end{itemize}

    Since the task-related methods require adaptation to produce the expected outputs, we describe each modified component to facilitate experimental replication, even if the code will be available on GitHub\footnote{Code available at \url{https://gitlab.liris.cnrs.fr/cifre_styx4d2024/seracfalldet}}:
    \begin{itemize}
      \item Dense feature matchers~\cite{LoFTR, ASpanFormer, TopicFM+} can generate change maps using only the initial part of their architectures. The complementary confidence from the coarse matches prior to refinement is interpolated to the original image resolution to obtain the final change map.
      \item For RoMa~\cite{RoMa}, the complementary certainty is extracted similarly.
      \item DepthAnythingv2~\cite{DepthAnythingv2} (relative depth) is used to assess whether current MDE models can effectively assist volumetric change detection. The absolute difference between two depth maps is computed, and depth differences exceeding twice the standard deviation are marked as changes, assuming large depth shifts correspond to meaningful scene variations.
    \end{itemize}
    
    In the generalization experiments, UniMatchv2~\cite{UniMatchv2} qualitatively demonstrated strong generalization in its pre-softmax activations, while its performance deteriorated when relying on its final softmax outputs. To illustrate this, we compared both outputs (pre- and post-softmax) for semi-supervised methods, considering a change when the pre-softmax activation exceeded two standard deviations above the mean.
    
    All evaluations were conducted using an NVIDIA RTX 3090 GPU with 24 GB of VRAM. Image pairs were processed patch-wise to handle memory limits, given the small size of events. Except for the previously stated modifications, both training and evaluation followed the original authors' implementations.

       \begin{table*}[!ht]
        \centering
        \captionsetup{justification=centering}
        \caption{\label{tab:zero-shot_comparison}Zero-shot comparison on SeracFallDet. Best (resp. second best) results in bold (resp. underlined).}
        \begin{tabular}[]{M{0.175\textwidth} M{0.135\textwidth} M{0.145\textwidth} M{0.110\textwidth} M{0.085 \textwidth} M{0.105\textwidth} M{0.065\textwidth}}
        \toprule
             Model & Task & Training dataset & Precision (\%) & Recall (\%) & F1 Score (\%) & IoU (\%) \\ \midrule
            LoFTR\textsuperscript{*}~\cite{LoFTR} & \multirow{4}{*}[-0.2em]{Matching} & \multirow{4}{*}[-0.2em]{MegaDepth~\cite{MegaDepth}} & 1.76 & \textbf{31.6} & 6.36 & 3.76 \\
            ASpanFormer\textsuperscript{*}~\cite{ASpanFormer} & & & \textbf{4.91} & 25.6 & \underline{8.33} & \textbf{5.07} \\
            TopicFM+\textsuperscript{*}~\cite{TopicFM+} & & & 1.37 & \underline{29.2} & 5.10 & 2.97 \\
            RoMa\textsuperscript{*}~\cite{RoMa} & & &  0.14 & 14.1 & 0.70 & 0.36 \\ \midrule
            DepthAnythingv2\textsuperscript{*}~\cite{DepthAnythingv2} & Monocular Depth Estimation & Depth-Anythingv2 data source~\cite{DepthAnythingv2} & 0.09 & 0.20 & 0.18 & 0.09 \\ \midrule \midrule
            SemiCD~\cite{SemiCD} & \multirow{8}{*}[-0.2em]{\shortstack{Change\\detection}} & \multirow{4}{*}[-0.2em]{\shortstack{LEVIR\_CD~\cite{LEVIR_CD}\\(10\%)}} & 0.00 & 0.00 & 0.03 & 0.01 \\
            SemiCD\textsuperscript{*}~\cite{SemiCD} & & & 0.11 & 0.50 & 0.46 & 0.25 \\
            UniMatchv2~\cite{UniMatchv2} & & & 0.00 & 0.00 & 0.00 & 0.00 \\
            UniMatchv2\textsuperscript{*}~\cite{UniMatchv2} & & & 0.23 & 8.39 & 3.15 & 1.70 \\ \cmidrule{0-0} \cmidrule{3-7}  
            SimSaC~\cite{SimSaC} & & \multirow{2}{*}[-0.2em]{VL-CMU-CD~\cite{VL-CMU-CD}} & 0.86 & 0.37 & 0.44 & 0.23 \\
            C-3PO~\cite{C-3PO} & & & 0.00 & 0.00 & 0.00 & 0.00 \\ \cmidrule{0-0} \cmidrule{3-7}  
            AnyChange~\cite{AnyChange} & & \multirow{2}{*}[-0.2em]{SAM-1B~\cite{SAM}} & 1.73 & 3.57 & 1.81 & 0.99 \\
            GeSCF~\cite{GeSCD} & & & \underline{2.30} & 17.4 & \textbf{8.52} & \underline{4.88} \\ \bottomrule
        \end{tabular}
        \caption*{* depicts models which have been slightly modified (Section \ref{section:baseline_methods}) to infer actual change.}
    \end{table*}

    \section{Results and discussions\label{section:results}}
    Based on the previously defined experimental setup, we evaluated the baseline models on SeracFallDet. This analysis, developed in the following section, leads us to an interpretation of the obtained results, providing insights into why a particular category of models appears to perform better in this context, as well as the limitations they are facing.
        
    \subsection{Generalization performances\label{section:model_evaluation}}
    Using the previously defined metrics, namely the pixel-wise IoU and F1 Score and event-wise recall and precision, we first compared different methods (Table \ref{tab:zero-shot_comparison}) in a zero-shot manner. This comparison might be biased by the contents of the various datasets on which the tested models have been trained. However, none of these datasets seems close to the evaluated dataset (Section \ref{section:dataset}).
    
    The performances of DepthAnythingv2~\cite{DepthAnythingv2} demonstrate that depth estimators do not trivially solve this task yet. \\
    The various matchers, all pretrained on MegaDepth~\cite{MegaDepth}, seem to generalize better on unknown scenes compared to change detectors. This might be due to the reverse change process that most of them follow, where they must provide sufficiently good descriptions (except RoMa~\cite{RoMa}, which is a regressor) for each pixel tile in both images so that similar tiles can be matched.
    Change detectors, on the other hand, have either a tendency to overfit on a specific kind of change to detect (e.g., building construction, object disappearance, etc.), or to focus on a similar task which might not be usable in every context (e.g., semantic segmentation) when supervised. This often leads to an empty change map. On the other hand, UniMatchv2~\cite{UniMatchv2}, a semi-supervised change detector, shows relevant generalization properties, particularly in its pre-classification output stage. This is especially noticeable, since this network was trained on significantly fewer labeled images compared to its fully supervised competitors. The performance obtained by GeSCF~\cite{GeSCD} is also remarkably good in its category. Despite the use of a segmentation model to perform change detection, a portion of its pipeline is similar to what is done with matchers~\cite{LoFTR, ASpanFormer, TopicFM+} by computing features similarity, which could explain its performance.
    
    \subsection{Qualitative results}
\begin{figure*}[ht]
        \centering
        \includegraphics[width = 0.95\textwidth]{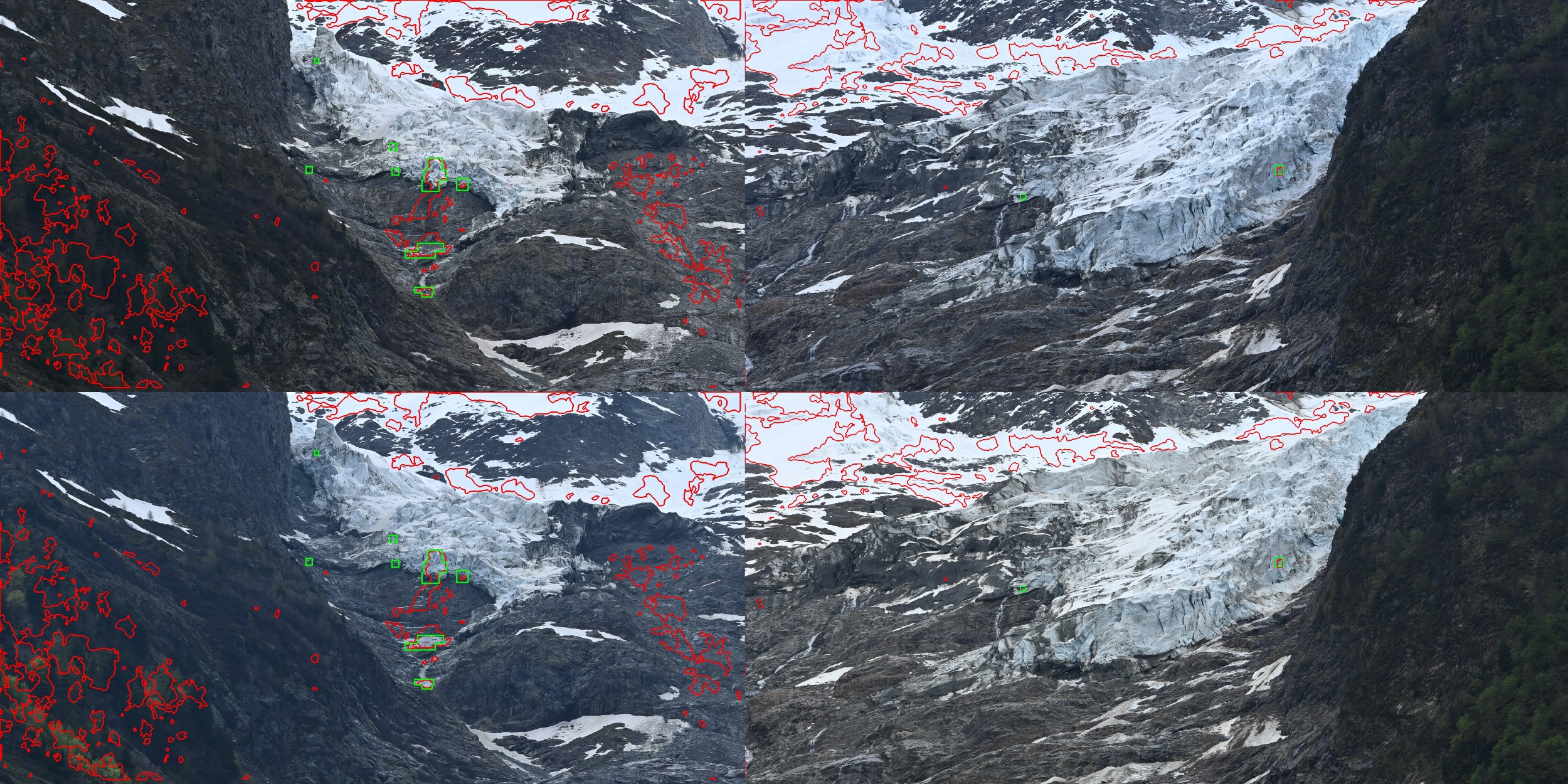}
        \caption{LoFTR~\cite{LoFTR} inference example on two image pairs (contours in red), before (first row) and after (second row) a serac fall (in green) \label{fig:qualitative_result}}
    \end{figure*}

    Even though the matchers demonstrate globally superior generalization on the SeracFallDet dataset, significant room for improvement remains, especially regarding the precision. The qualitative results (Figure \ref{fig:qualitative_result}) offer valuable insights on models' behavior. In fact, when handling texture-less tiles (typically sky or snow patches), matchers appear unable to pair them. This limitation is understandable, since each tile must possess a highly similar description, leading to potential matches with any comparable patch (sky or snow patch). \\
    Another condition degrading their performances happens when they are confronted to lighting variation with minimal texture. Since the models have not learnt extensively from highly variable lighting contexts, the described tiles might not be invariant to such conditions. However, this issue is mitigated when sufficient texture is available to guide the model's descriptions. \\       

    Moreover, detection becomes increasingly challenging as the event area shrinks.
    On a positive note, these methods exhibit the capability to associate image pairs characterized by inherent movements\textemdash such as glacier movement or misregistration. Although the matchers seem able to learn to detect changes without supervision, they are still facing significant challenges change detectors can ignore such as texture-less patches.
    
    \subsection{Supervised performances\label{section:trained_model_evaluation}}
    To demonstrate the challenges of this task for the current methods, we trained and evaluated some change detectors on the training split of SeracFallDet (Table \ref{tab:supervised_comparison}). For supervised models, only pairs of images with changes (Taconnaz 1 and Taconnaz 2 glaciers) were used, while the semi-supervised model UniMatchv2~\cite{UniMatchv2} used a mix of 40\% labeled and 60\% unlabeled pairs (within the training split). The unlabeled images were randomly selected with a maximum one-week gap between images in the pair.
    Overall, UniMatchv2, which uses a DINOv2 backbone and DPT decoder, outperformed C-3PO (based on VGG16 and DeepLabv3). This suggests that recent foundation models and vision transformers generalize better for this task. However, C-3PO achieved a particularly high recall score, likely due to the ability of CNNs to extract local features and detect smaller changes. 
    On the other hand, consistency regularization improved event-wise precision at the expense of a lower recall when leveraging unlabeled data. However, its effectiveness in learning more discriminative representations remains unclear, particularly given the lack of improvement in pixel-wise metrics (F1 score and IoU).

    Nonetheless, these results demonstrate that automatic serac fall detection remains an open challenge. Addressing it may require either further training with sufficient data to improve model reliability, or a shift in the training methodology. Future methods could benefit from combining the strengths of current change detectors with semi-dense features matchers, by learning in a self-supervised manner and improving the ability to disregard tiles similar to multiple others.
                
    \begin{table*}[ht]
        \centering
        \captionsetup{justification=centering}
        \begin{tabular}[]{ M{0.15\textwidth} M{0.15\textwidth} M{0.11\textwidth} M{0.09\textwidth} M{0.105\textwidth} M{0.07\textwidth}}
        \toprule
             Model & Architecture & Precision (\%) & Recall (\%) & F1 Score (\%) & IoU (\%) \\ \midrule
            UniMatchv2~\cite{UniMatchv2} (semi-supervised) & \multirow{2}{0.1\textwidth}[-0.4em]{\shortstack{DINOv2~\cite{DINOv2}\\\&\\DPT~\cite{DPT}}} & \textbf{31.1} & 27.7 & 31.4 & 20.9 \\ \cmidrule{0-0} \cmidrule{3-6}
            UniMatchv2~\cite{UniMatchv2} (supervised only) & & 25.2 & 32.8 & \textbf{33.3} & \textbf{22.3} \\  \midrule
            \shortstack{C-3PO~\cite{C-3PO}\\(supervised only)}& VGG16~\cite{VGG} \& DeepLabv3~\cite{DeepLabv3}  & 3.67 & \textbf{46.8} & 13.1 & 7.47 \\ \bottomrule
        \end{tabular}
        \caption{\label{tab:supervised_comparison}Training comparison on SeracFallDet. Best results are in bold.}
    \end{table*}
        
    \section{Conclusion and Perspectives\label{section:conclusion}}
        To conclude, current change detection methods are incorporating spatiotemporal coherence and understanding through feature interaction modules at the end of siamese encoders~\cite{FCNCD, DR-TANet, Changer, HANet, C-3PO}. These modules typically rely on either simple feature subtraction~\cite{FCNCD, C-3PO} or more complex, parameterized methods~\cite{DR-TANet, HANet}. While feature subtraction is computationally efficient, it may impose overly restrictive constraints on the feature space, forcing features to be artificially close. In contrast, feature similarity measures (e.g., dot product) could offer a more flexible approach, as they allow features to be aligned not only in magnitude but also in direction. They have naturally become a recent topic of interest~\cite{GeSCD, ZSSCD, AnyChange, SimSaC}. \\
        Recent advances have leveraged unlabeled data during training to achieve near-supervised performance levels. However, change detection datasets often exhibit a class imbalance toward the "no-change" class~\cite{LEVIR_CD, WHU_CD, VL-CMU-CD}. This imbalance complicates the learning process and can lead to feature collapse, where models tend to predict mostly empty change maps. Importantly, pairs of images without changes still contain valuable information that, when properly exploited, could enhance the learning process and mitigate these issues. \\
        The misregistration problem in change detection has been addressed using optical flow methods~\cite{SimSaC, SimSaCv2, sakurada_2017, Changer}. However, to our knowledge, no research has yet explored the effectiveness of these methods in the context of intra-frame motion, even though optical flow theoretically could solve, leaving an open question for future investigation. \\
        Despite the significant progress in change detection, some aspects have yet to be explored to enable generalization across different contexts (Table~\ref{tab:zero-shot_comparison}). Currently, this task has typically been approached as the appearance, disappearance, or replacement of whole objects~\cite{C-3PO, Sakurada_2020, AnyChange, MTP, ZSSCD, ZeroSCD}, limiting their applicability to scenarios where such assumptions do not hold (Section~\ref{section:dataset}). Given that datasets may not fully represent all possible change events, ensuring robustness across diverse and novel scenarios remains an unresolved issue. \\
        On the other hand, dense feature matchers, which aim at detecting as many correspondences as possible, appear to be antagonistic to traditional change detectors. While dense matchers excel in identifying correspondences, they struggle with texture-less patches and fail to distinguish between relevant changes and noise (e.g., variations due to lighting conditions). Change detectors, in contrast, are already designed to handle such cases. \\
        
        Future methods could benefit from combining the strengths of both approaches. For instance, integrating the robust classification of irrelevant changes from change detectors with the dense correspondence capabilities of feature matchers could lead to more accurate and generalizable models.

    \section*{Acknowledgement}
    We extend our sincere gratitude to all following contributors for providing access to the glacier footage that forms our dataset: Voltalia, DDT-74 and IGE, communauté de commune de la Vallée de Chamonix Mont-Blanc (CCVCMB) and COMPAGNIE DU MONT BLANC/Aiguille du Midi (www.montblancnaturalresort.com). Their contributions have been invaluable to this dataset. \\
    This work was supported by grant CIFRE number 2024/0695 from ANRT.

    % \printcredits
    \bibliographystyle{unsrt}
    \bibliography{SOTA}
    \addcontentsline{toc}{section}{References}

\end{document}